\documentclass[letterpaper, 10pt, conference]{ieeeconf}

\IEEEoverridecommandlockouts
\usepackage{cite}
\usepackage{amsmath,amssymb,amsfonts}
\usepackage{graphicx}
\usepackage{textcomp}
\begin{document}

\title{Key Findings for Small UAS for Post-Disaster Damage Assessment from the CURSE'26 Exercise}

\author{Thomas Manzini$^{1*}$, Stephen Johnson$^{1}$, Priyankari Perali$^{2}$, David Merrick$^{3}$, and Robin Murphy$^{1}$% <-this % stops a space
\thanks{$^{*}$Corresponding author.}% <-this % stops a space
\thanks{$^{1}$Thomas Manzini, Stephen Johnson, and Robin Murphy are with Texas A\&M University
        {\tt\small tmanzini@tamu.edu, sjohnson440@tamu.edu, robin.r.murphy@tamu.edu}}%
\thanks{$^{2}$Priyankari Perali is with the University of Maryland, College Park
        {\tt\small perali@umd.edu}}%
\thanks{$^{3}$David Merrick is with Florida State University
        {\tt\small dmerrick@em.fsu.edu}}%
}

\maketitle

\begin{abstract}

The CLARKE building and road damage assessment AI classifier was incorporated into the Catastrophic Unmanned Remote Sensing Exercise (CURSE) 2026, which trained 23 small uncrewed aerial systems (sUAS) teams from state and local agencies and non-governmental organizations in Florida and Missouri on how to collect imagery for an incident command Remote Sensing Cell. Although CLARKE has been operationally deployed to two hurricanes (Debby and Helene), CURSE'26 presented an opportunity to evaluate the machine learning model with a larger set of never-before-seen orthomosaics. The results show that building and road damage assessment together ran in 3.5 minutes on average, and that building damage assessment achieved a Micro F1 of 0.834 with no manual adjustments to correct spatial misalignment and delivered that performance in both Florida and Missouri, demonstrating good geographic transfer. It also showed that road damage assessment performed poorly due to sensitivity to misalignment between road maps and the images, and to lighting and environmental conditions in an urban setting. The analysis of CURSE'26 presents five other findings of value to the sUAS community: that the use of sUAS and machine learning is now operationally normalized and accepted, sUAS teams in the field can improve classification accuracy by timing their flights, exercises are good operational tests but their value for evaluating learning models is less clear, edge computing is needed to overcome the wireless upload barrier, and the application of sUAS and AI to disaster response has ethical and legal ramifications for pilots and data managers.

%CURSE'26 illustrates the value of exercises with high operational fidelity for robotics testing,

%but also highlights the ethical obligation of developers to provide human supervision of machine learning products. 

% was tested at the CURSE 2026 training exercise for NO PILOTS from NO agencies
%using unmanned aerial systems for disaster response. 
%As sUAS have become commonly used to rapidly collect high-resolution imagery before satellite or crew aviation imagery is available, a major barrier in tactical and strategic disaster response is the time to manually process the data. Imagery often takes 3-5 days for human experts to label damage. 
%CLARKE addresses this problem through CV/ML, architecture, and training on CRASAR-U-DROIDs.
%It was deployed on a small scale at Hurricanes Debby and Helene. 
%As part of training sUAS teams for disaster response, CLARKE was integrated into the CURSE 2026 exercise and evaluated. 
%The results show that building damage assessment ran in 5-10 minutes, achieved a Micro F1 of 0.834 with no adjustments, and delivered that performance in both Florida and Missouri. 
%It also showed that road damage assessment performed poorly due to sensitivity to road map-image misalignment and to lighting conditions in an urban setting. 
%CURSE'26 illustrates the value of exercises with high operational fidelity for robotics testing, but also highlights the ethical obligation of developers to provide human supervision of machine learning products. 

\end{abstract}


\section{Introduction}
\label{sec:intro}

\begin{figure}[t]
\centerline{\includegraphics[width=\columnwidth]{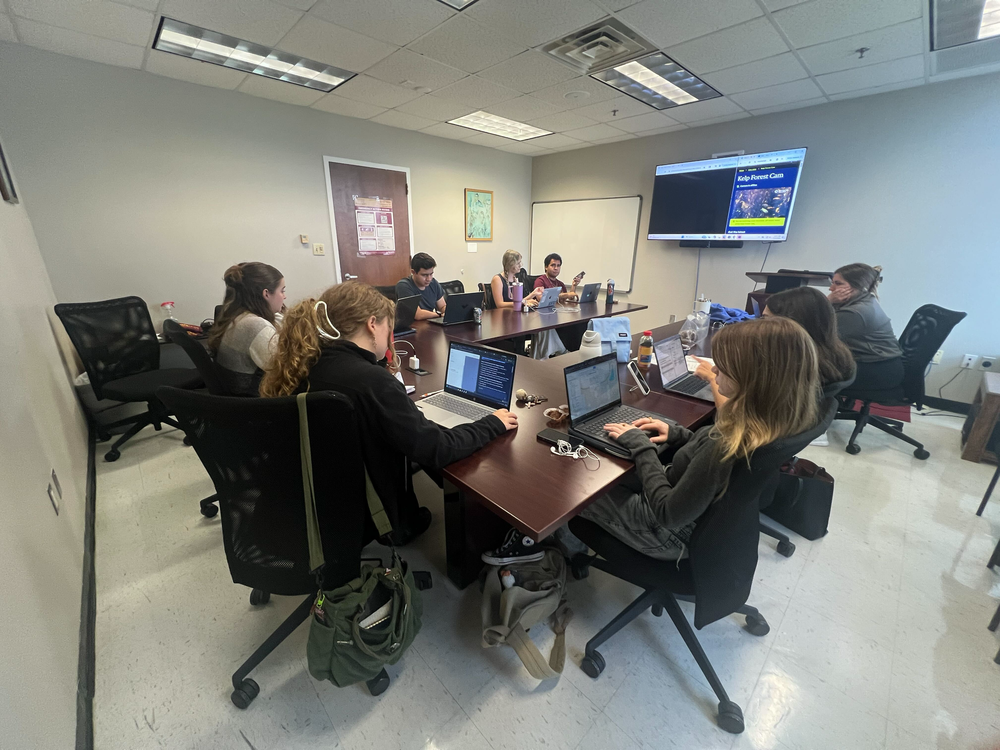}}
%\vspace{3.5in}
 \caption{The Remote Sensing Cell at CURSE'26 integrating sUAS imagery.}
\label{fig:pilots}
\end{figure}

On May 19, 2026, Florida State University's Emergency Management and Homeland Security (EMHS) program conducted the Catastrophic Unmanned Remote Sensing Exercise (CURSE) 2026. The full-scale exercise 
was designed to evaluate the use of small uncrewed aerial systems (sUAS) and remote sensing workflows during a catastrophic incident.
The 20 participating agencies and non-governmental organizations in Florida and Missouri operated from their home jurisdictions, flying sUAS to collect and process imagery locally, then transmitting the imagery and derived products to a centralized Remote Sensing Cell (RSC) located in Tallahassee (Fig.~\ref{fig:pilots}). The RSC integrated the data from the sUAS into a common operating picture and distributed products to stakeholders, completing the sUAS data-to-decision process as would be done during an actual disaster.

An integral part of the CURSE sUAS data-to-decision process was 
CLARKE (Computer vision and Learning for Analysis of Roads and Key Edifices), an automatic building and road damage assessment classifier.
CLARKE had been deployed during the operational responses to Hurricanes Debby and Helene in 2024 \cite{manzini2026deploying, manzini2025challenges} and was included in CURSE'26 as the ``new normal'' for producing initial damage assessment reports within minutes. 
CLARKE provides automated building and road damage assessment for sUAS-acquired orthomosaics. Building damage assessment is labeled according to the commonly used Joint Damage Scale (JDS) \cite{gupta2019creating}, while road damage uses a novel labeling schema as there are no existing standards. The building damage assessment component uses an encoder-decoder architecture consisting of a fine-tuned Vision Transformer \cite{reed2023scale} encoder and a UperNet \cite{xiao2018unified} decoder, and the road damage assessment component uses an Attention U-Net architecture \cite{oktay2018attention} trained on the CRASAR-U-DROIDs dataset \cite{manzini2024crasar}. This dataset ensures models have high operational fidelity because it contains over 67 gigapixels of imagery collected directly by responding agencies across 10 federally declared disasters.

%From  https://em.fsu.edu/curse26.php designed to evaluate the use of unmanned aircraft systems (UAS) and remote sensing workflows during a catastrophic incident. The exercise will be conducted on May 19, 2026, and is developed and conducted by the Emergency Management and Homeland Security (EMHS) Program at Florida State University in coordination with state, local, and federal partners.

%Unlike traditional co‑located exercises, CURSE'26 employs a statewide distributed model. Participating agencies operate from their home jurisdictions, conducting UAS data collection and processing missions locally, while transmitting imagery and derived products to a centralized Remote Sensing Cell (RSC) located in Tallahassee. The RSC serves as the primary fusion, analysis, and coordination hub, integrating data from multiple sources into a common operating picture, and distributing products to stakeholders.

%The exercise focuses on validating end‑to‑end remote sensing operations, including mission development and tasking, collaborative collection planning, resource management, data transfer, analytics, and product dissemination. CURSE'26 will utilize artificial intelligence algorithms developed by Texas A&M University and the NSF‑funded AI Institute for Societal Decision Making to produce initial damage assessment reports. These tools will be evaluated for their ability to accelerate analysis and improve situational awareness under realistic operational conditions.

%CURSE'26 integrated CLARKE into normative operations expected to be carried out by all Florida sUAS teams, rather than opportunistic insertion during a disaster. This was not the first time many of the sUAS teams at CURSE'26 had been exposed to CLARKE. 
%A mixture of N responders and emergency managers from M agencies had taken online awareness training classes on CLARKE. However, the 
%online course focused on the data products, not on how the sUAS teams would correctly acquire imagery that could be reliably processed by CLARKE or the methods for packaging and transferring the imagery to the CLARKE processor. 

The CURSE'26 exercise provided a mechanism to explore the
sUAS data-to-decision pipeline, rather than just the veracity of the CLARKE output. It confirmed sUAS and machine learning problems encountered at Debby and Helene \cite{manzini2025challenges} and Hurricane Ian \cite{manzini2023wireless},  
obtained positive feedback from real users, and identified new challenges and research needs.
CURSE'26 also presented an opportunity for examining whether exercises are realistic tests of sUAS and machine learning for sUAS-acquired imagery. 

This paper is organized as follows. A brief description of 
prior work provides background for comparing the exercise to previous research. 
Next, a description of the exercise follows, including background aimed at roboticists on how sUAS are situated within the larger data-to-decision process for damage assessment. The lessons learned for building damage assessment are presented, followed by the results from the road damage assessment. Three cross-cutting issues are discussed. The paper concludes with eight findings expected to be of value for the continued responsible innovation of sUAS systems for disaster management.

\section{Prior Work}
\label{sec:prior}

The CURSE'26 exercise differs from previously published field research on sUAS for damage assessment in general and CLARKE in particular.
At the time of writing, CLARKE is one of two known operationally deployed sUAS damage assessment systems \cite{manzini2025challenges, roman2025uav}. However, the integration of both systems into the sUAS data-to-decision process has been ad hoc and applied to only a small number of orthomosaics, with four used by CLARKE in the US \cite{manzini2025challenges} and four from the 2024 Valencia flash flooding in Spain \cite{roman2025uav}.
Consequently, the CURSE'26 exercise serves as a testbed offering a larger-scale deployment to systematically evaluate the formal operational pipeline.

%The CURSE'26 exercise provided a venue in which to confirm or deny expectations that i) orthomosaics would be processed in $\approx 5$ minutes, ii) transfer to a different geographical region may introduce problems, and iii) the wireless upload time, not the inference time, would be the major temporal bottleneck. 
%
\subsection{Automated Damage Assessment}

The research community predominantly assumes a seamless data-to-decision pipeline heavily reliant on wireless connectivity and cloud computing infrastructure \cite{qu2021dronecoconet, erdelj2017wireless, boccadoro2021extensive}. However, operational deployments have repeatedly demonstrated that wireless connectivity in post-disaster environments is either severely degraded or entirely unavailable, creating a critical bottleneck \cite{manzini2023wireless, manzini2025challenges, manzini2026deploying}. For instance, during the response to Hurricane Ian, the massive volume of high-resolution sUAS data generated in bursts by field teams consistently overwhelmed available cellular and satellite uplinks, forcing data managers to physically transport hard drives to intact emergency operations centers \cite{manzini2023wireless, manzini2023quantitative}. Because existing ML frameworks fail to account for these bandwidth constraints, an exercise such as CURSE'26 can highlight the distinct gap in understanding how to effectively execute inference at the edge during an active deployment. 
%The CURSE'26 exercise shows that even under normative, non-disaster training conditions, wireless upload times remain a bottleneck in the sUAS-to-decision workflow, indicating that ML inference at the edge is an operational necessity rather than a theoretical alternative.

\subsection{CLARKE}
\label{sec:CLARKE}

Previous efforts with processing sUAS imagery and developing and deploying CLARKE identified four issues: \textit{testing with a small set of never-before-seen orthomosaics} \cite{manzini2025challenges}, \textit{transfer to new geographical locations} \cite{manzini2025challenges}, \textit{wireless bandwidth limitations} \cite{manzini2023wireless}, and \textit{pilot error} \cite{murphy2024analysis, manzini2023wireless}. The deployments of CLARKE enabled evaluation of the first three, with pilot error out of the scope for feasible data collection.

The BDA and RDA classifiers that comprise the machine learning element were described in \cite{manzini2025challenges, manzini2026deploying},
along with how they were used at Hurricanes Debby and Helene \cite{manzini2025challenges}.
Those works show that 
CLARKE has been tested for only a small set of never-before-seen orthomosaics (4) at Debby and Helene \cite{manzini2025challenges}. 
Further, CLARKE showed weak transfer performance on imagery from Pennsylvania, where Hurricane Debby degenerated into its post-tropical remnants. This is
most likely due to the differences between the rural mountains of Pennsylvania and the flat coastal regions where hurricanes and tropical storms normally occur. The CRASAR-U-DROIDs dataset does not contain mountainous regions. 
Wireless bandwidth limitations that the sUAS pilots encountered with the data-to-decision process during the response to Hurricane Ian (2022) were discussed in \cite{manzini2023wireless}, where an expert manually classified the images. The CURSE'26 normative workflow is a codification of the Ian workflow, where CLARKE replaces the expert, so concerns about wireless limitations are still valid. 
% Pilot error at Hurricane Ian with the data-to-decision process was reported in \cite{murphy2024analysis, manzini2023wireless}, but is outside the scope of this paper. 

\section{Overview of sUAS Damage Assessment at CURSE'26} 
\label{sec:pipeline}

An overview of the sUAS Damage Assessment workflow at CURSE'26 consists of two parts. One is a description of the data-to-decision workflow that responders were trained on. The other is a summary of the nine CURSE'26 missions, which are compared with the Debby and Helene deployments in terms of geography, scale, land use, and type of damage. Fig.~\ref{fig:pipeline} presents the operational sUAS-to-damage assessment process for the nine sUAS damage assessment (DA) missions conducted in two states, captured in
Table~\ref{tab:missions}. The nine orthomosaics analyzed here were collected by seven teams.

\subsection{Pipeline from sUAS to Emergency Managers}

\begin{figure*}[t]
\centerline{\includegraphics[width=6.75truein]{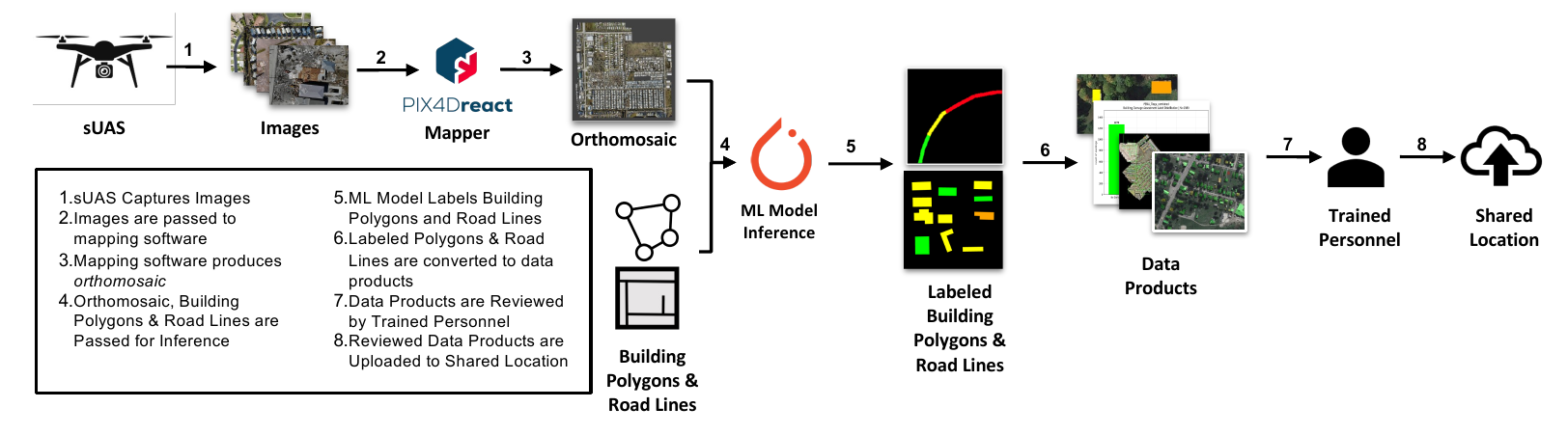}}
 \caption{The sUAS to emergency manager pipeline. The pipeline takes sUAS imagery and spatial data that are provided as input to the CLARKE system, providing DA for buildings and roads in the form of data products, which are reviewed by trained personnel and uploaded to a shared location for emergency managers.}
\label{fig:pipeline}
\end{figure*}

%pipeline
The pipeline diagram in Fig.~\ref{fig:pipeline} illustrates the workflow that CURSE'26 participants were trained on in generating actionable damage assessment information from sUAS imagery. It is a formalization of the process used at Debby and Helene.

Steps 1-3 are actions taken by the sUAS team. The team conducts an aerial survey, collecting raw imagery from a nadir perspective at a flight altitude determined by the pilot. The teams then create an orthomosaic map from the images using PIX4Dreact, a lightweight software package that can produce an orthomosaic reliably on a consumer laptop \cite{pix4dreact_misc}. Other programs could be used, but PIX4Dreact has become the standard of practice because of cost, speed, and reliability.
In step 4, the orthomosaics are uploaded by the teams to the data manager, either via wireless from the field or by transporting an external drive or SD card to the nearest base of operations, and are passed with the building polygons and road lines to the model for inference.

Steps 5-8 are where the data manager, one of the authors, applies the machine learning model to infer the damage categories of buildings and roads. The data manager does not know the sUAS or any details of the flights; they work only with the data. To reduce computational effort, the labeled damage is restricted to known building polygons and known roads. The pipeline uses Microsoft's US Building Footprints \cite{MicrosoftBuildingFootprints} for buildings and OpenStreetMap \cite{OpenStreetMap} roads, as they are commonly used in emergency management, are comprehensive, and are publicly available. If locally sourced building and road polygons are available, they could be readily leveraged as well.
The output of step 5 is labeled buildings and roads.  In order to accommodate the multiple possible uses and concurrent users, the output in step 6 is packaged into different outputs (e.g., a GIS-compatible file, a .png suitable for visual inspection, a spreadsheet of building addresses and damage, a map showing road conditions, etc.).

In step 7, the results are examined by the data manager for correctness before being distributed to emergency managers. This examination is necessary for two reasons. 
%1
The first reason is, as described in \cite{manzini2025non}, that ground maps and sUAS-generated maps have GPS errors, often leading to spatial misalignment of polygons and road lines. The CLARKE system does not correct the spatial misalignment of building polygons or roads. Fortunately, manual alignment, while not ideal in terms of speed and workforce, can be applied to both. The second reason is 
transfer of the ML model to new geographical locations and settings, or encountering new out-of-distribution disasters, as discussed in Sec.~\ref{sec:CLARKE}.
% The CLARKE system has learned to correct the spatial misalignment of building polygons but not roads. Fortunately, manual alignment, while not ideal in terms of speed and workforce, can be applied to roads. 
%2
%A second potential problem is the transfer of the ML model to new geographical locations and settings, or encountering new out-of-distribution disasters. The deployment of the system to Hurricane Debby for the beach areas of Florida performed well, but poorly for Tropical Debby in the rural mountains of Pennsylvania \cite{manzini2025challenges}. Given that each disaster is unique, transfer is a major concern during deployments. 

Finally, the workflow finishes with step 8, which sends the data products to the emergency managers at the RSC.  Neither the sUAS teams nor the data manager usually has any direct interaction with the emergency managers or each other; all three groups are working in silos. The emergency managers, as data consumers, expect the data products to be correct within their expectations.

\subsection{sUAS Damage Assessment Missions}
\label{sec:missions}

\begin{table}[t]
\caption{sUAS damage assessment missions at CURSE'26. GSD is the ground sample distance of each orthomosaic.}
%\centerline{\includegraphics[width=\columnwidth]{Figures/missions CURSES'26 corrected.pdf}}
\centering
\small
\setlength{\tabcolsep}{3pt}
\resizebox{\columnwidth}{!}{%
\begin{tabular}{|c|c|l|r|r|r|r|r|}
    \hline
    Mission & State & sUAS Model & \begin{tabular}[b]{@{}c@{}}GSD\\(cm/px)\end{tabular} & \begin{tabular}[b]{@{}c@{}}Orthomosaic\\Area (km$^2$)\end{tabular} & \begin{tabular}[b]{@{}c@{}}Number of\\Buildings\end{tabular} & \begin{tabular}[b]{@{}c@{}}Length of\\Roads (km)\end{tabular} & \begin{tabular}[b]{@{}c@{}}BDA+RDA\\Inference\\Time (s)\end{tabular} \\
    \hline
    1 & FL & Skydio X10 & 2.28 & 0.1048 & 2 & 0.24 & 214 \\
    2 & FL & Skydio X10 & 1.27 & 0.0469 & 1 & 0.70 & 188 \\
    3 & FL & Skydio X10 & 1.35 & 0.0261 & 0 & 0.25 & 139 \\
    4 & FL & WingtraOne & 5.00 & 0.2085 & 6 & 2.64 & 161 \\
    5 & FL & Skydio X10 & 3.26 & 0.5701 & 149 & 22.85 & 341 \\
    6 & FL & Skydio X10 & 2.16 & 0.0730 & 21 & 3.25 & 204 \\
    7 & FL & Skydio X10 & 1.59 & 0.1204 & 24 & 1.52 & 235 \\
    8 & MO & DJI Matrice 30 & 4.02 & 0.1587 & 19 & 1.75 & 181 \\
    9 & MO & DJI Matrice 30 & 12.70 & 1.8807 & 98 & 17.31 & 214 \\
    \hline
\end{tabular}
}
\label{tab:missions}
\end{table}

The differences between the nine orthomosaics and those generated at Debby and Helene are:
\begin{itemize}
    \item \textit{Geographic diversity.} Seven of the nine missions were conducted at different sites in Florida, with the remaining two in Missouri. This replicates the geographical diversity experienced at Debby, which spanned Florida and Pennsylvania.
    
\item \textit{Scale.} The average orthomosaic at CURSE'26 covered 0.354 km$^2$, half the size of the average orthomosaic at Debby and Helene, which was
0.792 km$^2$ \cite{manzini2025challenges}.
The CURSE'26 orthomosaics were also smaller than the average of 1.307 km$^2$ collected at Hurricane Ian \cite{manzini2024crasar}, the largest known deployment of sUAS at a disaster \cite{manzini2023wireless, manzini2023quantitative}. As indicated by the number of buildings, the mission locations ranged from rural (0 buildings in mission 3) to high-density urban (149 buildings in mission 5). The average road length was 5.61 km.

\item \textit{Land use.} Because these nine missions were conducted during non-disaster training exercises, they provide a unique opportunity to evaluate the models against new land uses. To avoid any public concern over safety or privacy, CURSE'26 missions were generally tasked to fly over industrial parks, urban facilities, or underpopulated rural areas rather than targeting residential neighborhoods, which were a high priority during Debby and Helene. 

\item \textit{Types of damage.} The missions also provide a unique opportunity to evaluate the models against a near-zero damage baseline. 
A manual assessment of all captured imagery found only 31 instances of building damage out of 320 buildings, or 9.7\% of buildings. For roads, the baseline physical road damage was estimated to be $\approx5\%$, primarily due to roads obscured by vegetation and construction.
\end{itemize}

Although a direct comparison between the inference times at CURSE'26 listed in Table~\ref{tab:missions} and Debby and Helene is not possible, the fast inference time at CURSE'26 is expected to be applicable to deployments using laptop computers. The average inference time using a desktop computer with an NVIDIA GeForce RTX 4090 (24 GB) GPU and an Intel Core Ultra 9 285K (3.70 GHz) CPU with 128 GB RAM was 209 seconds (3.5 minutes). Such hardware is reasonably deployable at a base of operations, and even though the CURSE'26 orthomosaics were smaller, it is expected that equivalent inference times can be achieved on laptops with similar NVIDIA GPUs.

%\section{Building Damage Assessment with UAS Findings}
\section{Building Damage Assessment}
\label{sec:BDA}

\begin{figure}[t]
    \centering
    \includegraphics[width=\columnwidth]{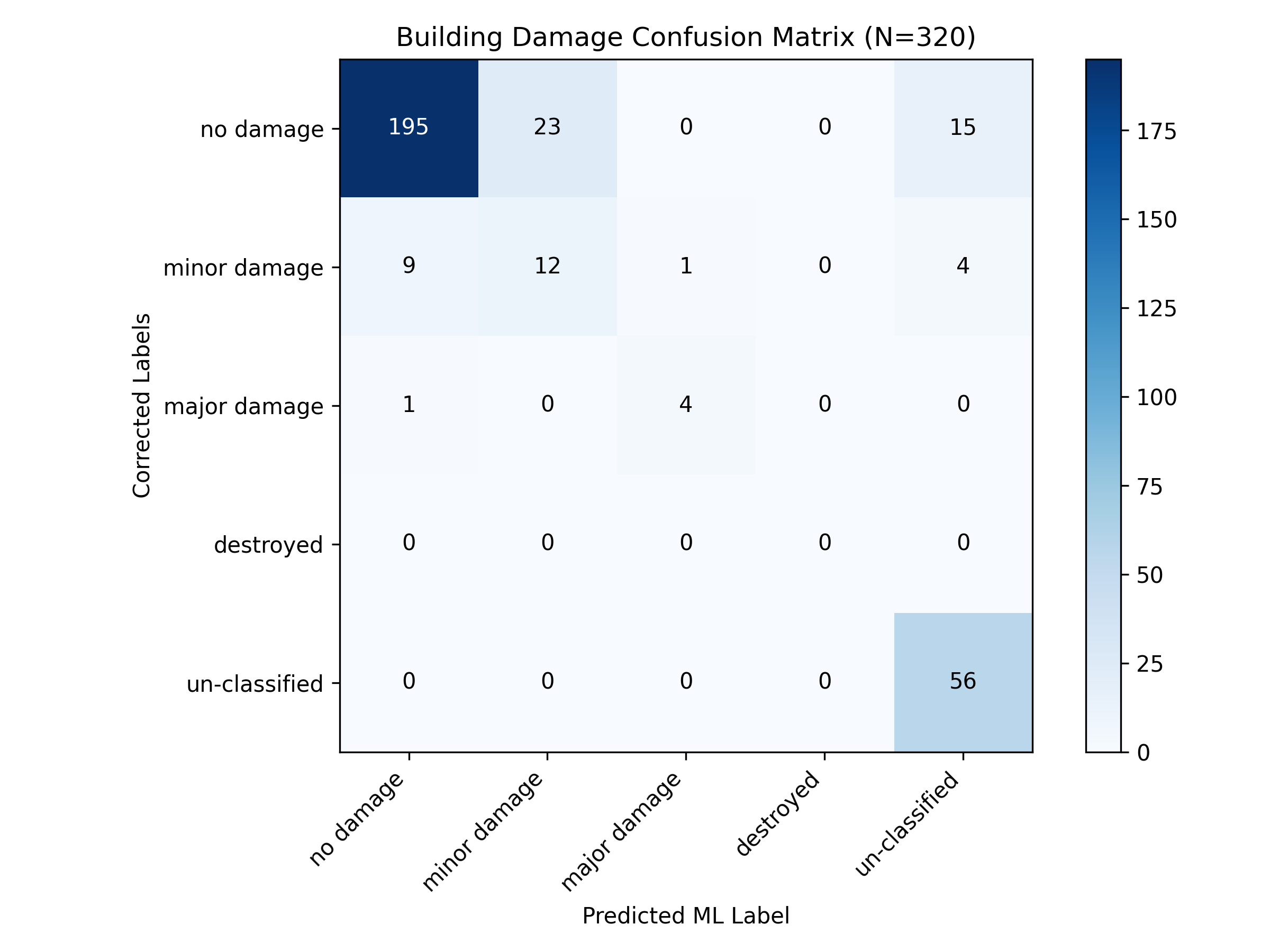}
    \caption{The confusion matrix for all building damage assessment missions.}
    \label{fig:confusion-matrix}
\end{figure}

The Building Damage Assessment (BDA) component functioned well for each of the eight missions that captured buildings, measured in terms of F1, transfer, and lack of need for manual adjustments to correct spatial misalignment. It should be noted that as this event was an exercise, the system was not run on actual disaster imagery, meaning that the system performance on the complete collection of JDS classes could not be evaluated. However, the exercise did highlight three potentially significant failure modes, particularly in
commercial and larger buildings. Of these failure modes, only poor illumination can be mitigated by the sUAS; pilots should try to fly as close to noon as possible, though the operations tempo of a real disaster may not permit this. 

\subsection{Performance}

While fully evaluating the correctness of model predictions is not possible given the incomplete and otherwise highly unbalanced class distributions, Fig.~\ref{fig:confusion-matrix} shows the full confusion matrix and that the performance of the model was acceptable on the CURSE'26 orthomosaics, as it presents a Micro F1 of 0.834. The ground truth labels were collected by the authors who have direct expertise responding to major disasters and curating building damage assessment labels for use in machine learning models \cite{manzini2024crasar}. The reader should note the heavy class imbalance towards the ``no damage'' class, and that prior work has evaluated these models on CRASAR-U-DROIDs using Macro F1 \cite{manzini2026deploying}. This limited class distribution forces an evaluation using micro-average F1.

In addition to the reasonable F1 score, the model showed positive transfer to the orthomosaics from Missouri. Given the amount of hurricane imagery from the state of Florida in the CRASAR-U-DROIDs training set, it was not surprising that the model performed well on the six Florida missions. It was surprising to see that the model transferred to the two Missouri missions, given the different terrain and the problems encountered with the deployment of the model during Hurricane Debby. Though claims of complete transfer cannot be made as this exercise did not permit the evaluation of the model on the complete set of JDS classes, there was no obvious or persistent failure across the imagery. Notably, the most similar data in the CRASAR-U-DROIDs dataset, the 2021 Mayfield, Kentucky, tornado disaster, was included in the test set and was not used for model training.
From a practical workflow aspect, the BDA component did not require any manual adjustments to the spatial alignment of building polygons.

\subsection{Failure Modes}

\begin{figure}[t]
\centerline{\includegraphics[width=\columnwidth]{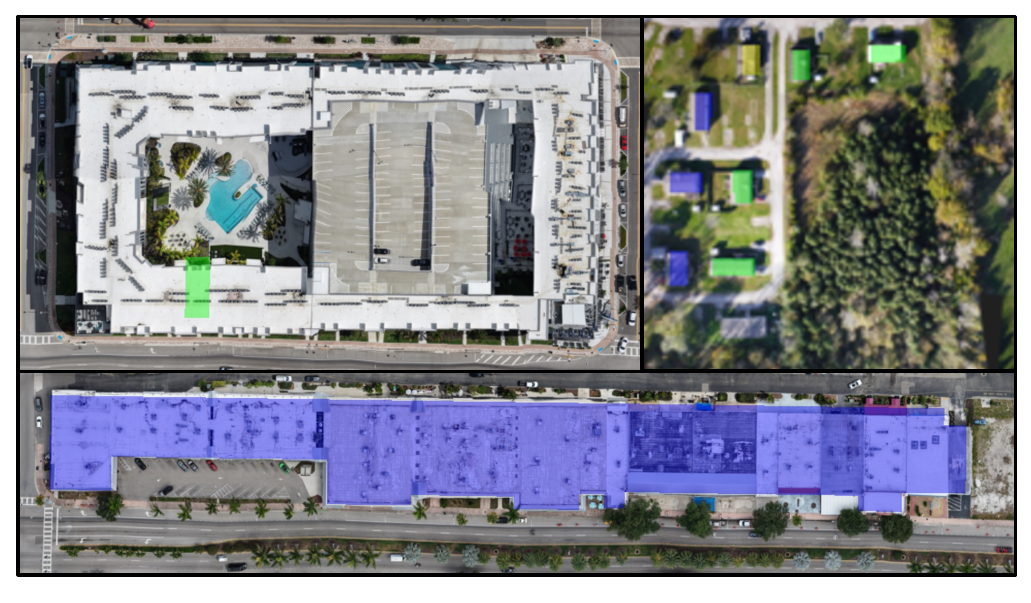}}
 \caption{Three examples of failure modes for the BDA model. [Upper-Left] Example of an incorrect spatial polygon for a building. [Upper-Right] Example of incorrect building damage classification of the left side buildings affected by illumination factors. [Bottom] Example of incorrect building damage classification for a large building (predicted ``un-classified'' vs. actual label ``no damage''). Classes are represented by the following colors: (green, ``no damage''), (yellow, ``minor damage''), (blue, ``un-classified'').}
\label{fig:bda-failures}
\end{figure}

 Fig.~\ref{fig:bda-failures} shows three different cases of failures: incorrect building polygons, under-representation of large structures, and illumination issues. 
The large apartment complex in the upper left is represented by a single green polygon in Microsoft's US Building Footprints \cite{MicrosoftBuildingFootprints}. This is a failure of the footprint data, not of the model.
The polygon for the long, narrow manufacturing facility at the bottom is correct, but the model classifies it as ``un-classified'' despite no obvious damage, possibly because large commercial and multi-unit residential structures are under-represented in the CRASAR-U-DROIDs dataset \cite{manzini2024crasar}. 
The upper right image again shows instances where a building is mislabeled as ``un-classified'' likely due to illumination factors (e.g., shadow casts) affecting the buildings' visual representations. Those images were taken in early morning or late afternoon rather than the desired midday lighting.

\section{Road Damage Assessment}
\label{sec:RDA}

Based on the qualitative performance during the CURSE'26 exercise, the Road Damage Assessment (RDA) component of the CLARKE system is not ready for operational deployment. It labeled 35.3\% of the road length across the nine orthomosaics as damaged against an expected baseline of approximately 5\%, with one orthomosaic showing nearly 80\% ``total damage''. The poor performance of the RDA component appears to be due to three failure modes present in the imagery. These three are discussed in detail below.

%due to its reliance on manual interventions and its sensitivity to low-relevance illumination factors (e.g., shadows). As detailed in the subsequent performance evaluation, applying the RDA baseline across the nine missions required substantive manual spatial adjustments to align road lines correctly with the orthomosaics. The focus of this section is on identifying specific operational barriers that demonstrate that the current RDA pipeline requires a level of human intervention that precludes fully automated use.

\subsection{Performance}

\begin{table}[t]
  \caption{Road damage per mission. Note: All missions should have reported 95\% or higher values for ``no damage''.}
  %\centerline{\includegraphics[width=\columnwidth]{Figures/RDADamageSummary corrected.pdf}}
  \centering
  \small
  \resizebox{\columnwidth}{!}{%
  \begin{tabular}{|c|r|r|r|r|}
      \hline
      Mission & \begin{tabular}[b]{@{}c@{}}Length of\\Roads (km)\end{tabular} & \begin{tabular}[b]{@{}c@{}}No Damage\\(\%)\end{tabular} & \begin{tabular}[b]{@{}c@{}}Partial Damage\\(\%)\end{tabular} & \begin{tabular}[b]{@{}c@{}}Total Damage\\(\%)\end{tabular} \\
      \hline
      1 & 0.24 & 74.5 & 0.0 & 25.5 \\
      2 & 0.70 & 85.2 & 6.5 & 8.3 \\
      3 & 0.25 & 15.3 & 5.0 & 79.7 \\
      4 & 2.64 & 85.7 & 0.1 & 14.2 \\
      5 & 22.85 & 49.9 & 11.4 & 38.7 \\
      6 & 3.25 & 34.0 & 17.0 & 49.0 \\
      7 & 1.52 & 61.7 & 0.6 & 37.7 \\
      8 & 1.75 & 89.5 & 2.1 & 8.4 \\
      9 & 17.31 & 84.1 & 0.9 & 15.0 \\
      \hline
  \end{tabular}
  }
  \label{tab:rda_damage_amounts}
\end{table}

%\begin{table*}[htbp]
%  \caption{Damage Label Count per Orthomosaic for Road Damage Assessment}
%  \centerline{\includegraphics[width=2\columnwidth]{Figures/RDA Label Counts.png}}
%  \label{tab:rda_label_counts}
%\end{table*}

%\begin{table*}
%  \centering
%  \caption{Percent of Total Label Count per Damage Tier per Orthomosaic for Road Damage Assessment provided by CLARKE}
%  \centerline{\includegraphics[width=2\columnwidth]{Figures/RDA Label Percentages.png}}
%  \label{tab:rda_label_percentages}
%\end{table*}

The performance of the RDA component was poor for all orthomosaics. 
As shown in Table~\ref{tab:rda_damage_amounts},
the nine orthomosaics contained a total of 50.52 km of roads, ranging from 0.24 to 22.85 km per image. For example, in Table~\ref{tab:rda_damage_amounts}, the orthomosaic from mission 3 labeled nearly 80\% of the roads as ``total damage''. None of the missions were labeled with $\approx 95\%$ of the roads as ``no damage'', as would be expected from the manual inspection of the imagery. Given the obvious disparity between reality and predictions, generating a confusion matrix was not considered necessary.

%The baseline performance of the RDA classifier across the 9 missions reveals that applying the model directly to the 9 orthomosaics without human correction leads to widespread misclassifications throughout the road networks. The resulting uncorrected outputs, including the baseline kilometers of damage, raw label counts, and percentage distributions across the 9 missions, are quantified in Tables \ref{tab:rda_damage_amounts}, and \ref{tab:rda_label_percentages}. 

%Although the RDA data lacks exhaustive manual labeling for a formal numerical evaluation, the generated classifications for total and partial damage are demonstrably incorrect by a substantive margin. As established, the intact nature of these training environments dictates an expected physical damage baseline of $\approx5\%$ of all roads. Despite these intact conditions, the model's outputs show a clear tendency toward false-positive damage predictions. For instance, 1 of the 9 orthomosaics resulted in a classification where 58.9\% of the roads were labeled as experiencing total damage, as shown in Table \ref{tab:rda_label_percentages}. This stark contrast between the zero-damage reality and the elevated damage predictions directs the conclusion that the baseline classifier cannot currently produce reliable metrics without human oversight.

\subsection{Failure Modes}

\begin{figure}[t]
\centerline{\includegraphics[width=\columnwidth]{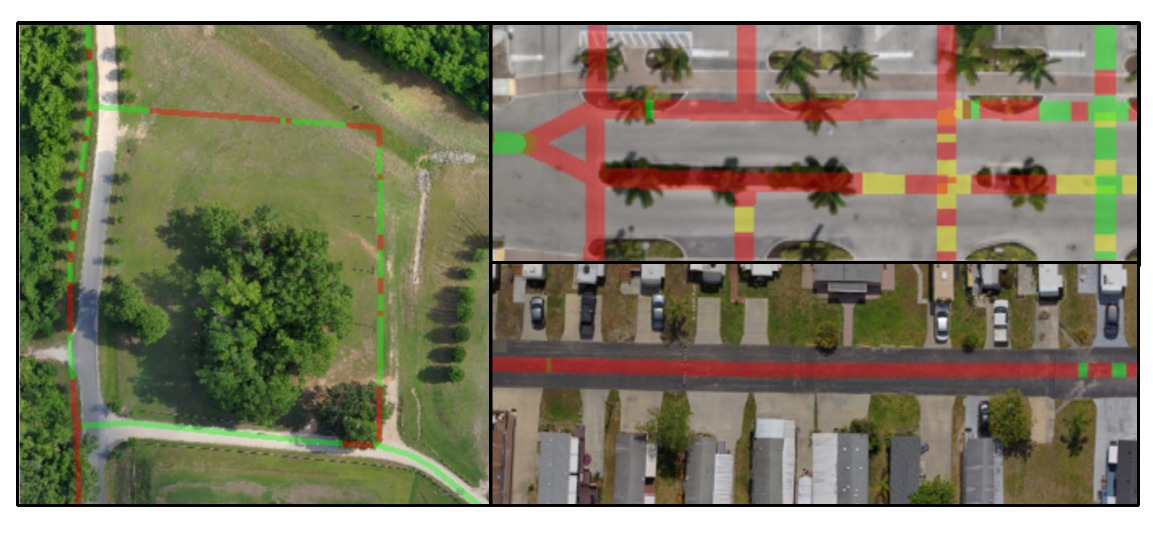}}
 \caption{Three examples of failure modes for the RDA model. Road damage labels are represented as green (``no damage''), yellow (``partial damage''), and red (``total damage''). [Left] Example of RDA misclassifications due to misaligned, incorrect spatial road lines, and illumination factors. [Upper-Right] Example of failure mode with road medians. [Bottom-Right] Example of RDA misclassifications with dark road color (tar roads).}
\label{fig:rda-failures}
\end{figure}

Three primary failure modes for the CLARKE RDA component were identified: spatial misalignment, unfavorable illumination and road color (e.g., shadows, tar roads), and environmental obscurations (e.g., trees). Of the three, spatial misalignment appears to have had the most significant impact on mislabeling. Only unfavorable illumination can be addressed by sUAS pilots, favoring flying midday or overcast conditions. Spatial alignment during post-processing could be improved by integrating a dedicated alignment model, following \cite{zampieri2018multimodal}. The environmental obscurations are a property of the real world and cannot be addressed by pilots or software directly. However, future efforts should be considered to better manage prediction uncertainty when faced with these types of obscurations.

%Inspection of the RDA classifier's outputs across the nine missions reveals two primary failure modes: spatial misalignment and attachment to low-relevance illumination factors and environmental features, such as shadows and trees, along roads. The primary finding of this evaluation is that the current RDA model requires substantive architectural adjustments and retraining to handle the spatial and visual diversity of non-ideal operational environments.

The RDA component exhibits a sensitivity to the spatial alignment of road lines, reflecting a vulnerability previously documented in the literature \cite{manzini2025non}. Specifically, the underlying training set expects the structural centerline of the spatial road line to rest exactly in the center of the road within the imagery \cite{manzini2026benchmark}. However, the operational version of CLARKE routinely encounters non-uniform spatial alignments when utilizing the publicly available a priori road maps (Steps 5 to 8 in Fig.~\ref{fig:pipeline}).
As shown in the upper right panel of Fig.~\ref{fig:rda-failures}, spatial misalignments, especially with roads with medians, direct the classifier to adjacent off-road terrain, generating false damage labels.

The RDA model also appears sensitive to illumination and road material variations. The left panel of Fig.~\ref{fig:rda-failures} illustrates a failure mode where shadows from landscaping elements direct the model to falsely predict ``total damage'' (red) on an intact surface. This effect is also present in the upper-right panel, where a central median with landscaping, coupled with spatial misalignment, leads the model to predict ``total damage'' (red) and ``partial damage'' (yellow) throughout the image. Furthermore, the bottom-right panel of Fig.~\ref{fig:rda-failures} demonstrates the RDA model's apparent response to dark road surface colors as it predicts ``total damage'' (red) on an otherwise clear road.

%Although the barriers posed by spatial misalignments and visual distractors are substantial, they can be mitigated through targeted data collection and pre-processing interventions. To mitigate visual misdirection caused by shadows, flight operations should be encouraged during overcast conditions or at solar noon, thereby limiting the presence of high-contrast visual artifacts. Furthermore, resolving the spatial misalignment requires either a manual correction of the road networks prior to inference or the integration of a dedicated alignment model \cite{zampieri2018multimodal} to automatically reconcile the expected centerlines with the collected imagery.
        
\section{Discussion}
\label{sec:discussion}

%The CURSE'26 exercise confirmed edge inference is an operational requirement to overcome wireless transmission bottlenecks. Furthermore, while the exercise proved valuable for validating the normative sUAS-to-decision workflow and ethical requirement for human oversight, the absence of naturalistic damage prohibits a full evaluation of the underlying machine learning capabilities.

The CURSE'26 exercise provides broader insights into the design and deployment of sUAS and artificial intelligence. The first insight is the complex role exercises can play in evaluating data-to-decision systems involving sUAS. The second insight is that even under ideal conditions, edge computing is necessary. The third insight is that sUAS deployments engender ethical concerns for pilots and data managers.

\subsection{Role of Exercises in Evaluating sUAS and ML}

CURSE'26 raises the issue of the role of real-world exercises in evaluating sUAS and artificial intelligence systems such as CLARKE for disasters. Are exercises operational tests of the pipeline but not good inference tests, or can they be both? CURSE'26 suggests that exercises are strong tests of the operational pipeline but weak tests of inference on the damage seen to date, although they are a useful source of edge cases that operational data cannot easily supply.

The exercise was a realistic test of how CLARKE performed as part of the operational workflow, but it was not a realistic test of the ML model. The imagery collected by the sUAS teams was not representative of the types of damage observed by actual sUAS deployments. The CURSE'26 imagery contained no disaster damage, and only 31 of the 320 buildings showed any damage at all. Missions were also often in commercial areas, versus the residential or rural areas typically targeted after an event. 

Even though the exercise was not a good test of the ML model for typical damage, CURSE'26 illustrates how an exercise can be a source for identifying or testing for future conditions. As described in Sec.~\ref{sec:missions}, the imagery of undamaged commercial or rural areas did not represent the types of buildings, roads, or damage seen to date in real-world disasters. 
Consider that the current practice is to send sUAS to a relatively small set of areas where there is expected to be damage based on hazard prediction maps or the emergency managers' knowledge of the community. As such, an entire orthomosaic with no damage is an important edge case that does not exist in the CRASAR-U-DROIDs dataset, which contains naturalistic disaster damage.

\subsection{Need for Edge Computing}

The exercise confirms prior expectations that edge computing of inference rather than reliance on cloud computing is needed to speed up the sUAS-to-decision pipeline.
The pipeline in Fig.~\ref{fig:pipeline} suggests three ways the process could be executed faster: faster data acquisition by the sUAS teams (step 1), post-processing (steps 2-7), and upload to a shared location thus enabling human decision-making (step 8). It is not clear that sUAS teams can fly faster, though there may be improvements in assigning missions. 
This leaves the steps of getting the imagery to the data manager and then processing it. The generation of orthomosaics is already done on the edge by the sUAS teams using commercial software optimized for speed. The inference time was on the order of 3.5 minutes. However, the upload times for the orthomosaics to the data manager were on the order of 1 hour, consistent with Hurricane Ian, though specific upload times were not recorded.
It should be noted that CURSE'26 was conducted under ideal conditions for communications, as the wireless infrastructure was intact, unlike in a real disaster. Therefore, it is expected that, based on the findings detailed in \cite{manzini2023wireless, manzini2023quantitative}, the biggest gain in decision speed with sUAS imagery is running inference at the edge.

\subsection{Responsible Innovation}

%1
Although CLARKE is autonomous for BDA, it requires human oversight of the data products. Such oversight is needed to ensure that there are no major degradations in accuracy due to different disasters,  geographies, or other out-of-distribution situations. The human oversight is not a casual glance at the output; it requires a data manager skilled in machine learning.

It should be noted that applying CLARKE without human oversight may be a form of negligent misrepresentation, because emergency managers form expectations about accuracy that the developers have neither corrected nor restricted, and thus developers may be legally liable for harms \cite{hu2026introduction}.
The sUAS pilots and data manager should be aware that it is their responsibility to ensure that their outputs are of the highest quality and do not show any of the failure modes in Secs.~\ref{sec:BDA} and~\ref{sec:RDA} because how the damage assessment products will be used and by whom is unknown. As shown in Fig.~\ref{fig:pipeline} step 8, the data products are uploaded to a common store in the RSC for access potentially by anyone in the incident command enterprise. If the data products are not within expectations, the decision-maker would not be aware of the problems.
Although running at the edge is desirable, it may not be an easy solution given the current state of automatic classification, the potential for errors to negatively affect the response, and the developers' legal liability.
%2

\section{Findings}

The examination of the CURSE'26 exercise in the context of previous deployments and research experimentation leads to eight findings. These findings contribute to disaster robotics research and practitioner communities:
\begin{enumerate}
    \item \textit{The use of sUAS with computer vision/machine learning for automated damage assessment is now operationally normalized and accepted.} It was incorporated into training 23 sUAS teams from two states at CURSE'26 as standard operating procedures for the upcoming hurricane season. The participants expressed satisfaction with CLARKE outputs.  

 \item \textit{sUAS teams in the field can improve CLARKE automated damage assessment by flying in overcast or midday conditions to minimize shadows.} As described in Sec.~\ref{sec:BDA} and Sec.~\ref{sec:RDA}, illumination has a noticeable impact on labeling damage. While the operational tempo of a response may not permit teams to choose when to fly, if they do have a choice, they should strive for the conditions that produce the highest image quality. 
 
     \item \textit{The use of sUAS with CLARKE for building damage assessment is mature.} 
    %component functioned well for each of the nine missions in terms of F1, transfer, and no need for
%manual adjustments. 
As described in Sec.~\ref{sec:BDA}, CLARKE performed well in terms of F1, transfer, and lack of need for manual adjustments to correct spatial misalignment.
CURSE'26 addressed the issue of being exposed to a small set of
never-before-seen images raised in Sec.~\ref{sec:CLARKE}. The exercise increased the number of new orthomosaics from 4 to 13.  
Although the CURSE'26 orthomosaics were smaller than previous events, the average inference time of 3.5 minutes given in Sec.~\ref{sec:pipeline} is consistent with the 5-7 minute times on the CRASAR-U-DROIDs dataset test set orthomosaics \cite{manzini2024crasar}. 

    \item \textit{CLARKE BDA should perform well in novel geographies.} CLARKE demonstrated zero-shot transferability to a novel geography, addressing the second issue raised in Sec.~\ref{sec:CLARKE}. As described in Sec.~\ref{sec:BDA}, CLARKE performed well in terms of F1 score, demonstrating strong positive transfer on the Missouri missions (8 and 9). This geographical generalization is particularly notable because similar Midwestern terrain (specifically the Mayfield, Kentucky, tornado imagery) in the CRASAR-U-DROIDs dataset was not used for training the models but was instead isolated for testing. This indicates that the BDA model can successfully generalize to out-of-distribution terrains, such as Missouri. %, while delivering 3.5-minute average inference times as long as there is good spatial alignment between the building map and the imagery. 

 \item \textit{The use of sUAS with CLARKE for road damage assessment (RDA) is not mature.} Unlike the BDA component, the RDA classifier misclassified portions of undamaged roads as having some form of damage; in one case, nearly 80\% of the roads were mislabeled. The primary culprit appears to be the model's high sensitivity to non-uniform spatial alignment errors between the road lines and the orthomosaics.
 %Consequently, the RDA system currently requires substantial, and potentially prohibitive, manual intervention to center and correct spatial misalignments before inference, limiting its use as a fully automated, rapid-response tool.

     \item \textit{Exercises are good operational tests, but their value in evaluating machine learning is less clear.}  CURSE'26 introduced imagery with two deviations from what has been collected from actual deployments. The first was the general lack of damage in the imagery, representing an inconsistency with respect to actual disaster imagery. The second was the prevalence of commercial buildings and urban areas. As described in Sec.~\ref{sec:discussion}, this means CURSE'26 was not realistic in that it was not consistent with imagery seen in the past, but it could be realistic as the scope of sUAS deployments expands. 
     
     \item \textit{Edge computing with the sUAS teams is needed to overcome the wireless upload barrier.} CURSE'26 confirms the concern over wireless bandwidth raised in Sec.~\ref{sec:CLARKE}. Although specific quantitative upload times were not recorded during the CURSE'26 exercise, it should be clear from Fig.~\ref{fig:pipeline} that transferring the data from the field to a data manager in a base of operations or emergency operations center introduces notable time delays, even for a one-day exercise. %This is consistent with what was observed at Hurricane Ian \cite{manzini2023wireless}.
     
    \item \textit{The application of sUAS and AI to disaster response merits close attention to the principles of responsible innovation.} Technologists must be aware of the ways in which their systems may fail when used for a disaster event and proactively prevent misunderstandings of accuracy or propagation of poor results. Human oversight is absolutely critical, and the introduction of faulty data products into an operational data-to-decision workflow could lead to negligent misrepresentation and induce meaningful harms. Technologists and operators must bear the ethical responsibility to verify model outputs and prevent the propagation of failures during operations.
    
%    \item \textit{CLARKE demonstrates zero-shot transferability to Missouri}. As described in Sec. \ref{sec:BDA}, CLARKE performed well in terms of F1 score, demonstrating strong positive transfer on the Missouri missions. This geographical generalization is particularly notable because similar midwestern terrain—specifically the Mayfield, Kentucky, tornado imagery—was strictly isolated in the CRASAR-U-DROIDs test set. This indicates that the BDA model can successfully generalize to out-of-distribution terrains, such as Missouri, while delivering 3.5-minute average inference times at the edge.
\end{enumerate}

CLARKE BDA is positioned to be used during the 2026 hurricane cycle by Florida and will continue to be refined. The findings show a strong need for edge computing by field teams, but at the same time argue for human supervision. Future work will improve the RDA component and explore automatic uncertainty quantification, so that the system itself can flag out-of-distribution situations.

\section*{Acknowledgment}
This work is supported by the AI Research Institutes Program funded by the National Science Foundation under the AI Institute for Societal Decision Making (NSF AI-SDM). The authors thank the FL-UAS1 task force,  Florida State University, and the organizers of the CURSE'26 Exercise for their support.

\bibliographystyle{IEEEtran}
\bibliography{conference_101719}

\end{document}